%% file: main.tex
\documentclass[runningheads]{llncs}

\usepackage{eccv}

\usepackage{eccvabbrv}

\usepackage{graphicx}
\usepackage{booktabs}
\usepackage{multirow}
\usepackage{tikz}

\usepackage[accsupp]{axessibility}  

\usepackage[pagebackref,breaklinks,colorlinks,citecolor=eccvblue]{hyperref}

\usepackage{orcidlink}

\newcommand{\ours}{\textsc{AnyImageNav}}

\newcommand{\sconf}{\mathcal{S}_\text{conf}}
\newcommand{\srel}{\mathcal{S}_\text{relev}}

\begin{document}

\title{\ours: Any-View Geometry for Precise Last-Meter Image-Goal Navigation} 

\titlerunning{Abbreviated paper title}

\author{Yijie Deng\thanks{Equal contribution.}\enspace$^{1,2,3,4}$, Shuaihang Yuan$^\ast$$^{1,2,4}$, Yi Fang\thanks{Corresponding author: Yi Fang yfang@nyu.edu.} \enspace$^{1,2,3,4}$\\}

\authorrunning{Y.~Deng et al.}

\institute{NYUAD Center for Artificial Intelligence and Robotics (CAIR), Abu Dhabi, UAE \and
New York University Abu Dhabi, Electrical Engineering, Abu Dhabi 129188, UAE \and
New York University, Electrical \& Computer Engineering Dept., Brooklyn, NY 11201, USA. \and
Embodied AI and Robotics (AIR) Lab, NYU Abu Dhabi, UAE.}
\maketitle

\begin{figure}[h]
    \centering
    \vspace{-20pt}
    \includegraphics[width=0.8\linewidth]{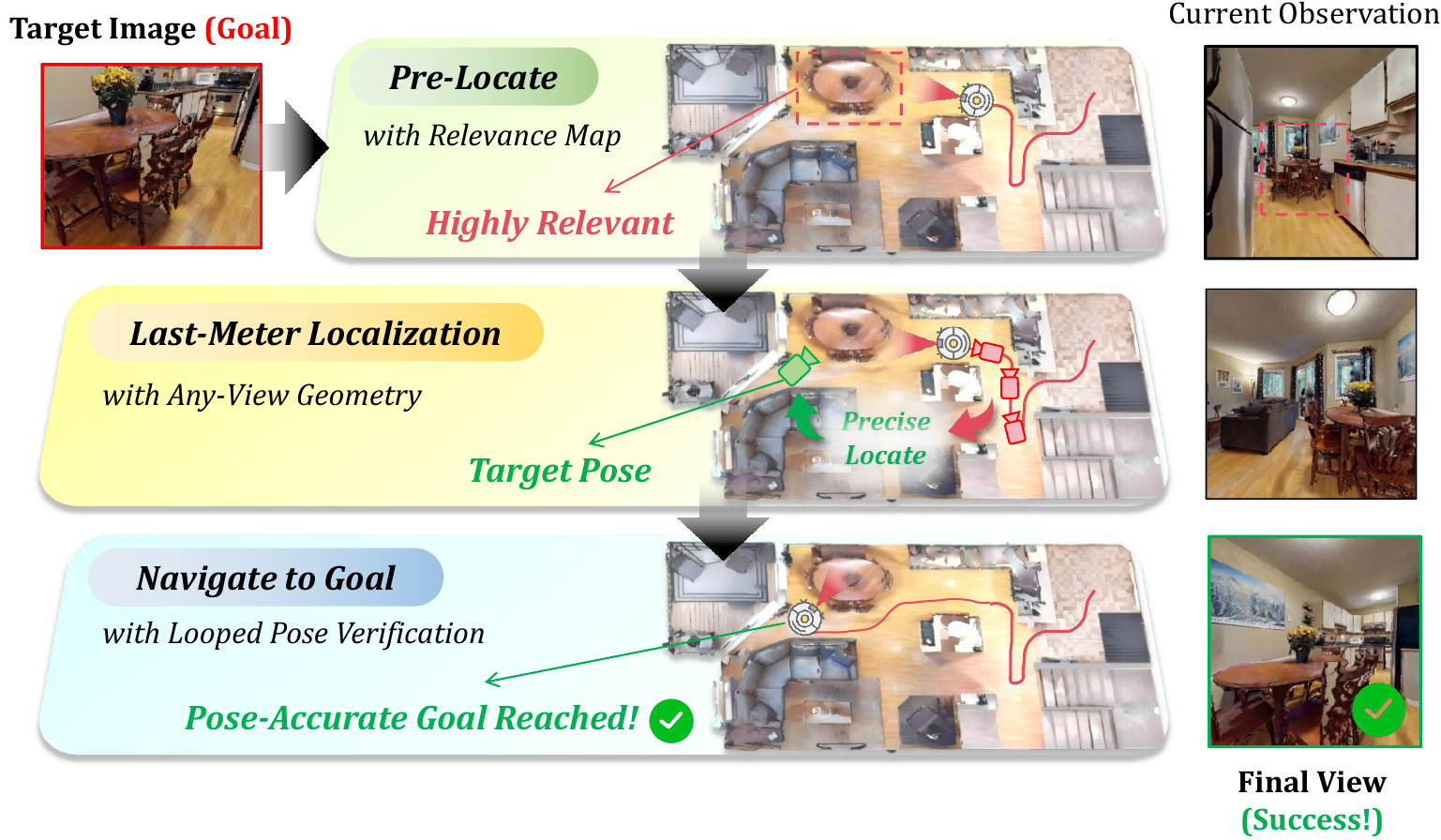}
    \vspace{-8pt}
    \caption{\ours{} overview. A BEV relevance map identifies highly relevant regions and guides the agent toward the goal vicinity (\textit{Pre-Locate}). Once a highly relevant region is detected, any-view geometry registers the goal image with previous observations to recover a precise target pose (\textit{Last-Meter Localization}). The agent then navigates to the pose that closely matches the goal viewpoint (\textit{Navigate to Goal}).}
    \vspace{-34pt}
    \label{fig:teaser}
\end{figure}

\input{subtex/0_abstract}
\input{subtex/1_introduction}
\input{subtex/2_related_work}
\input{subtex/3_method.tex}
\input{subtex/4_experiment}
\input{subtex/5_discussion}
\input{subtex/6_conclusion}

%
%
\bibliographystyle{splncs04}
\bibliography{main}
\end{document}

%% file: subtex/0_abstract.tex
\begin{abstract}
Image Goal Navigation (ImageNav) is evaluated by a coarse success
criterion---the agent must stop within 1\,m of the target, which is
sufficient for finding objects but falls short for downstream tasks such as grasping that require precise positioning.
We introduce \ours{}, a training-free system that pushes ImageNav toward
this more demanding setting.
Our key insight is that the goal image can be treated as a \emph{geometric query}: any photo of an object, a hallway, or a room corner
can be registered to the agent's observations via dense pixel-level
correspondences, enabling recovery of the exact 6-DoF camera pose.
Our method realizes this through a semantic-to-geometric cascade: a semantic
relevance signal guides exploration and acts as a proximity gate,
invoking a 3D multi-view foundation model only when the current view is
highly relevant to the goal image; the model then self-certifies its
registration in a loop for an accurate
recovered pose.
Our method sets state-of-the-art navigation success rates on Gibson (\textbf{93.1\%}) and HM3D (\textbf{82.6\%}), and achieves pose recovery that prior methods do not provide: a position error of \textbf{0.27\,m} and heading error of \textbf{3.41°} on Gibson, and \textbf{0.21\,m} / \textbf{1.23°} on HM3D, a \textbf{5–10$\times$} improvement over adapted baselines. Our project website: \href{https://yijie21.github.io/ain/}{https://yijie21.github.io/ain/}
\keywords{Image Goal Navigation \and 6-DoF Pose Recovery \and
Any-View Geometry \and Training-Free Navigation}
\end{abstract}

%% file: subtex/1_introduction.tex
\section{Introduction}
\label{sec:intro}

Consider a household robot given a reference photo of an object on a
shelf, asked to retrieve it.
Modern Image Goal Navigation agents can reliably bring a robot to
within 1\,m of a visual goal. But when that goal is a reference
photo for a downstream manipulation task, proximity alone is not
enough, because the robot must also know precisely \emph{where} and
\emph{from which direction} the photo was taken to act on what it sees.
This \emph{last-meter gap} is not merely a precision shortcoming; it
is the boundary between navigation and action, between finding and
doing.
Closing it requires the agent to recover the exact 6-DoF camera pose
of the goal image, not just its approximate vicinity.

Image Goal Navigation, including general image goal
navigation~\cite{7989381} and instance image goal
navigation~\cite{krantz2023navigating}, asks an agent to reach a
target viewpoint in an unknown environment guided only by a single
photograph.
Recent modular and end-to-end methods have pushed success rates to
impressive levels across diverse indoor
environments~\cite{lei2024instance,krantz2023navigating,yin2025unigoal,
narasimhan2025splatsearch,deng2025hierarchical,lei2025gaussnav,
al2022zero,chaplot2020neural,majumdar2022zson,sun2023fgprompt}.
However, today's success criterion is coarse: the agent must stop
within 1.0\,m of the goal, with the goal being oracle-visible by
turning or looking up and down.
This is sensible for navigation, since being within arm's reach and
having the target in view is meaningful.
Yet for downstream manipulation, this last meter is critical.
A household robot asked to grasp an object depicted in a reference
photo must reach the \emph{exact} position where the photo was taken
and face the object correctly.
We therefore emphasize the last-meter problem of image goal navigation
and seek to recover accurate 6-DoF poses, not merely proximity.

We identify two structural reasons why current methods cannot close
this gap.
\emph{Modular methods}~\cite{lei2024instance,lei2025gaussnav,
narasimhan2025splatsearch} compare rendered or sampled viewpoints to
the goal image and stop when a semantic similarity score exceeds a
threshold.
Because they reason about appearance rather than geometry, they can
determine that the agent is \emph{near} the goal but not precisely
\emph{where} or \emph{how} the goal camera was oriented.
The closest prior work to our aim,
GauScoreMap~\cite{deng2025hierarchical}, does reason about 6-DoF pose
but requires a pre-built Gaussian Splatting representation of the
environment, a costly, environment-specific reconstruction step that
must be repeated for every new scene.
\emph{End-to-end methods}~\cite{majumdar2022zson,sun2023fgprompt}
learn direct observation-to-action mappings; during early exploration,
when the current view shares little visual content with the goal, the
learned signal degrades and the policy must rely on blind exploration.
At a deeper level, both families treat localization as a
\emph{semantic} problem: they rely on recognizable object categories,
visual landmarks, or learned appearance embeddings.
This works coarsely, but semantics cannot recover an accurate camera
pose.

This paper asks: \emph{what if we treat the goal image as a geometric
query rather than a semantic one once the agent is in close
proximity?}
Modern 3D multi-view foundation models~\cite{wang2025vggt,wang2025pi,
lin2025depth} discover dense pixel-level correspondences across
arbitrary image collections and recover relative camera poses in a
single forward pass, with no scene reconstruction, no object
recognition, and no environment-specific training.
Crucially, these models operate at a finer granularity than the
category-level matching used in prior navigation
methods~\cite{lei2024instance,lei2025gaussnav,deng2025hierarchical},
and they have only recently matured to the reliability needed for
autonomous navigation commitment.
We show that this capability is the right primitive for precise
viewpoint recovery, bridging the gap between coarse proximity
navigation and manipulation-ready localization.

To fill this gap, we propose \ours{}, a training-free system
that treats any goal image as a geometric query, with the overview shown in Figure~\ref{fig:teaser}.
Its core is a \emph{semantic-to-geometric cascade} that unifies
exploration and localization around a single shared representation.
A pixel-level relevance signal between the current observation and the
goal image is computed at every step; it serves both as a frontier
scoring cue and as a proximity sensor that gates the 3D foundation
model.
When triggered, the foundation model self-certifies its registration
confidence from its internal cross-frame features before the agent
commits to the estimated 6-DoF pose, turning the model's intrinsic
correspondence quality into a navigation decision signal.

Our contributions are as follows:

\begin{itemize}

    \item \textbf{Pose-precision Image Goal Navigation.}
    We extend image goal navigation to a stricter precision regime and introduce
    a complementary evaluation protocol measuring position and
    heading direction errors, exposing the quantitative gap between
    proximity-based success and manipulation-ready localization that
    the standard criterion conceals.

    \item \textbf{Any-view geometric correspondence for navigation.}
    We show that the internal across-frame correspondence confidence, an incidental byproduct of the 3D multi-view foundation model's inference, can be directly repurposed as a navigation commitment signal: it reliably indicates whether the agent has entered the visual neighborhood of \emph{any} goal image, without semantic labels, object categories, or additional learned classifiers.

    \item \textbf{State-of-the-art results on both tasks.}
    \ours{} achieves state-of-the-art navigation success rates on both
    benchmarks: \textbf{93.1\%} on Gibson for general image goal
    navigation and \textbf{82.6\%} on HM3D for instance image goal
    navigation.
    It simultaneously delivers 6-DoF pose recovery that prior methods
    do not provide, achieving a position error of \textbf{0.27\,m}
    and a heading error of \textbf{3.41°} on Gibson, and
    \textbf{0.21\,m} / \textbf{1.23°} on HM3D, a 5--10$\times$
    improvement over adapted baselines and establishing the first
    strong baseline for precision-oriented image goal navigation.

\end{itemize}

%% file: subtex/2_related_work.tex
\section{Related Work}
\label{sec:related}

\subsection{Image-Goal and Instance-Image-Goal Navigation}

Image Goal Navigation (ImageNav)~\cite{zhu2017target,7989381} requires an
agent to navigate to the viewpoint from which a goal photograph was taken.
Instance-Image Goal Navigation (IIN)~\cite{krantz2022instance} narrows
this to images that depict a specific object instance, requiring the agent
to both find and discriminate among visually similar objects.
This paper addresses both settings within a unified framework.

\textbf{End-to-end methods} train a policy to map visual observations
directly to actions.
Early work~\cite{zhu2017target,al2022zero,yadav2023ovrl,yadav2023offline}
relied on reinforcement or imitation learning with CNN or recurrent
encoders.
FGPrompt~\cite{sun2023fgprompt} conditions the policy on fine-grained
goal prompts to improve goal-directed attention, and
RegNav~\cite{li2025regnav} incorporates region-level representations to
improve goal grounding.
While effective at coarse navigation, end-to-end methods produce no
explicit goal pose estimate and provide no principled mechanism for
precise last-meter localization: the policy's implicit representation of
goal proximity degrades as the observation diverges from the goal image,
which is common outside the immediate goal vicinity.

\textbf{Modular methods} decompose the pipeline into mapping,
exploration, and goal-matching stages.
Topological methods~\cite{savinov2018semi,kim2023topological} construct
graph-based memory structures for long-range navigation.
Renderable memory~\cite{kwon2023renderable} synthesizes novel viewpoints
to bridge the appearance gap between observation and goal.
Wasserman~\etal~\cite{wasserman2023last} specifically target the last-meter
problem by refining position estimates through careful feature matching,
though they do not recover full 6-DoF poses.
IGL-Nav~\cite{guo2025igl} and SplatSearch~\cite{narasimhan2025splatsearch}
leverage Gaussian Splatting for view synthesis and render-and-compare
matching.
For IIN, IEVE~\cite{lei2024instance} proposes an
exploration-verification-exploitation framework, and
UniGoal~\cite{yin2025unigoal} unifies multiple goal-conditioned navigation
tasks under a single zero-shot policy.
Despite strong success rates, all these methods stop when a semantic
similarity score exceeds a threshold, providing only approximate
positioning---they cannot answer \emph{where exactly} the goal camera was
pointing, which is the question \ours{} is designed to answer.

\textbf{Gaussian Splatting-based methods} represent the scene as a
collection of 3D Gaussians~\cite{kerbl20233d} to support textured
render-and-compare matching.
GaussNav~\cite{lei2025gaussnav} builds a semantic Gaussian map during a
dedicated exploration episode and matches rendered views to the goal image
in subsequent episodes.
GauScoreMap~\cite{deng2025hierarchical} extends this to hierarchical
scoring, using CLIP-derived relevance fields for coarse candidate selection
and local Gaussian geometry for fine pose estimation.
While GauScoreMap can in principle output a precise 6-DoF pose, both
methods require either a \emph{pre-built} scene representation that must
be completed before any navigation begins, or a dedicated first-pass
reconstruction episode.
This is a fundamental limitation: constructing a full Gaussian scene scales
poorly to large environments and is inapplicable in single-episode
settings.
\ours{}, by contrast, requires no pre-built representation and recovers
precise poses from the observations gathered during the navigation episode
itself, using a single forward pass of a 3D multi-view foundation model.

\subsection{Visual Localization and Pose Estimation}

Place recognition~\cite{arandjelovic2016netvlad} and visual
localization~\cite{sattler2018benchmarking} are classical computer vision
tasks closely related to our localization stage.
Hierarchical localization~\cite{sarlin2019coarse} chains a coarse retrieval
stage using global descriptors with a fine geometric verification stage
using local feature matching, a design that achieves state-of-the-art
accuracy on large-scale benchmarks.
\ours{} follows an analogous two-stage design within a navigation loop: the
semantic proximity score $\srel$ provides coarse screening, and the 3D
multi-view foundation model provides fine geometric registration, preventing
expensive geometric computation during the bulk of exploration.

Local feature matching methods~\cite{sarlin2020superglue,sun2021loftr}
establish dense correspondences between image pairs and are used in
classical localization pipelines as the fine-matching stage.
Our approach goes further: rather than matching the goal image to a
database of pre-mapped images, we register it directly against the agent's
accumulated observations in a single feed-forward pass, without a
pre-built scene database or iterative optimization.

\subsection{3D Multi-View Foundation Models}

Recent work has produced feed-forward models that jointly estimate camera
poses and 3D structure from unordered image collections.
DUSt3R~\cite{wang2024dust3r} introduced the paradigm of treating pairwise
3D reconstruction as a regression problem solvable by a transformer.
VGGT~\cite{wang2025vggt} extends this to sets of images with a
single-pass multi-view architecture that outputs per-pixel depth, camera
poses, and point clouds simultaneously.
Pi3~\cite{wang2025pi} and Depth-Anything-3~\cite{lin2025depth} further
scale this paradigm with larger training sets and stronger generalisation
to in-the-wild images.
RobustVGGT~\cite{han2025emergent} extends VGGT with an outlier-filtering
mechanism based on cross-image correspondence confidence, which \ours{}
repurposes as a navigation commitment signal.

These models were designed for offline 3D reconstruction tasks, not for
embodied navigation.
\ours{} is the first to exploit their capability as an \emph{online}
localization primitive: by treating the goal image as an unposed view to be
registered against the agent's keyframe history, we inherit their
correspondence-level precision without requiring scene reconstruction,
pre-built maps, or semantic recognition of the goal content.
This makes our approach the only ImageNav method that achieves precise
6-DoF goal pose recovery without any scene-specific preprocessing.

%% file: subtex/3_method.tex
\section{Method}
\label{sec:method}

\begin{figure}[t]
    \centering
    \includegraphics[width=\linewidth]{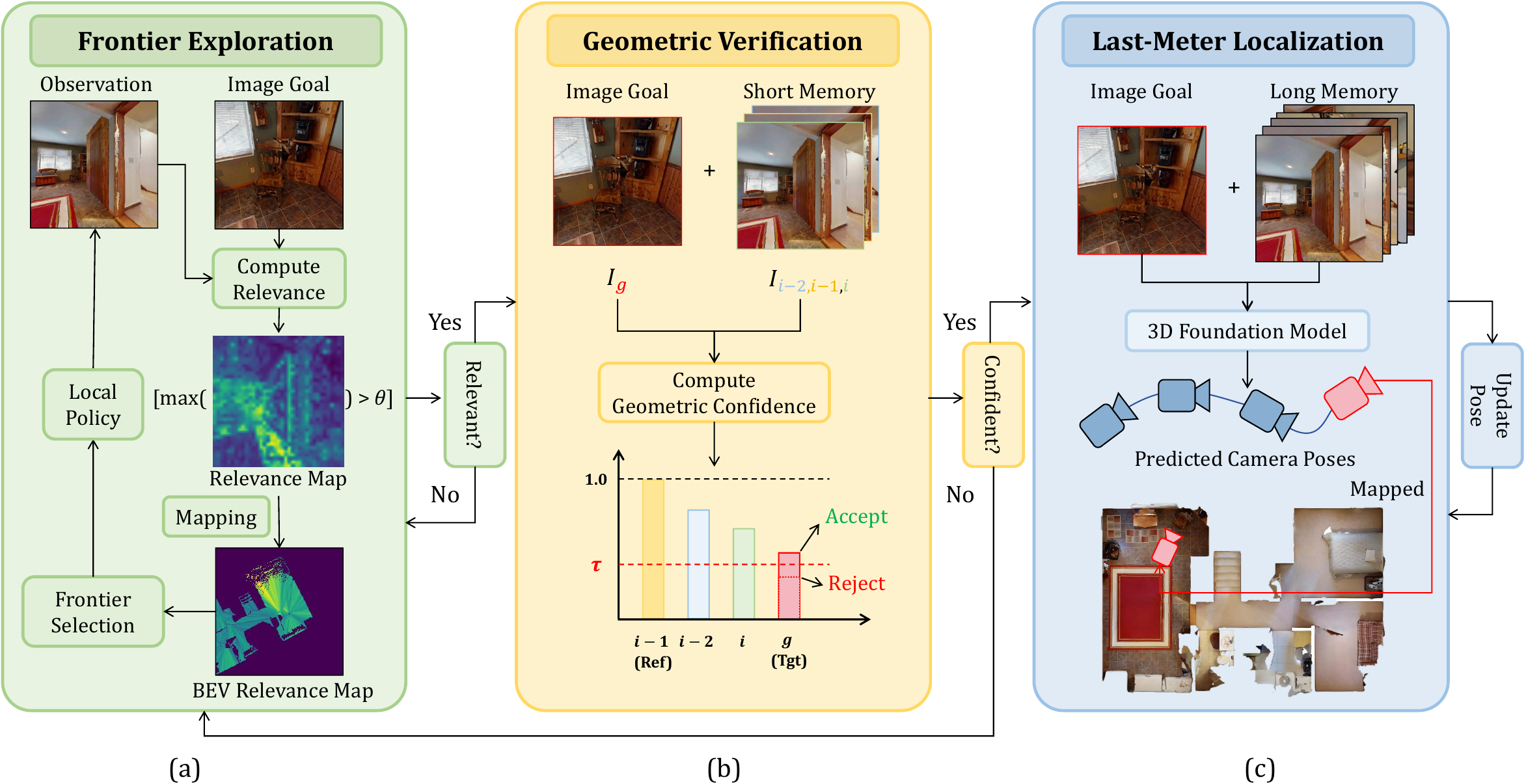}
    \caption{
        \textbf{\ours{} pipeline.}
        \textbf{(a) Frontier Exploration:} The dense relevance map
        between the current observation and the goal image $I_g$ is
        projected onto a BEV map, scoring candidate frontiers and
        yielding the proximity score $\srel$.
        \textbf{(b) Geometric Verification:} When $\srel > \theta$,
        a short memory window and $I_g$ are passed to a 3D multi-view foundation model; the
        model's internal correspondence confidence $\sconf$ gates
        the transition to localization.
        \textbf{(c) Last-Meter Localization:} Upon acceptance,
        the full long-memory cache is registered with $I_g$ in a
        single forward pass; the recovered goal pose is aligned to
        the agent's map via Sim(3)+orientation correction and
        continuously refined as confidence improves.
    }
    \label{fig:pipeline}
\end{figure}

\subsection{Task Setting}

At each timestep $t$ the agent receives an RGB image $I_t$, a depth
map $D_t$, and its camera-to-world pose $\mathbf{T}_t \in SE(3)$,
forming the observation $o_t = (I_t, D_t, \mathbf{T}_t)$.
The agent must navigate to the target image $I_g$ taken from an
unknown pose $\mathbf{T}_g$ in the same environment.
Beyond the standard proximity criterion
$\|\hat{\mathbf{p}} - \mathbf{p}_g\|_2 < 1\,\text{m}$, we
introduce two pose-precision metrics:
\begin{align}
    \epsilon_\text{pos} &= \|\hat{\mathbf{p}} - \mathbf{p}_g\|_2,
    \qquad
    \epsilon_\text{head} = |\hat{\psi} - \psi_g|_{180},
\end{align}
where $\hat{\mathbf{p}}, \mathbf{p}_g \in \mathbb{R}^2$ are the
2D positions of the agent's final pose and the goal pose projected
onto the ground plane, and $\hat{\psi}, \psi_g \in [0°, 360°)$
are the corresponding yaw angles;
$|\cdot|_{180}$ is the minimum angular difference modulo $180°$.

\begin{figure}[t]
    \centering
    \includegraphics[width=\linewidth]{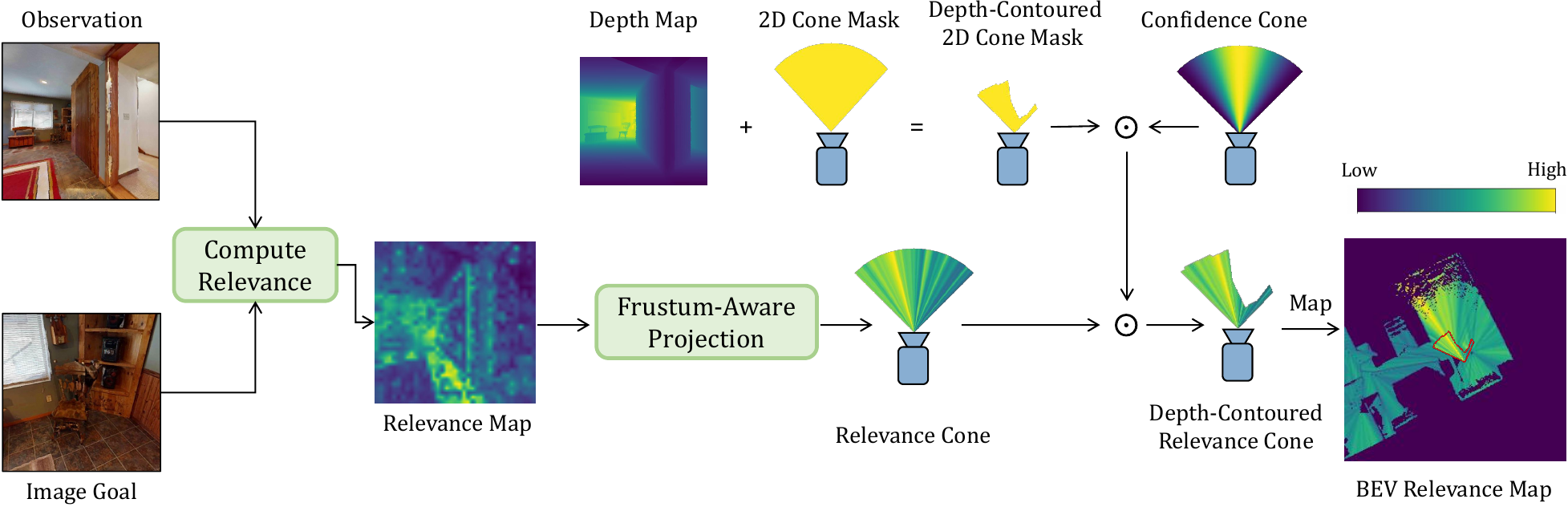}
    \caption{
        \textbf{BEV Relevance Map Construction.}
        A dense pixel-level relevance map between the current observation
    and the goal image is calculated per step.
    Then a frustum-aware projection then maps the relevance map onto the
    top-down BEV grid: a 2D cone mask is combined with depth-contoured
    truncation to prevent relevance from bleeding through obstacles,
    and element-wise multiplied with a confidence cone that
    attenuates rays toward the field-of-view boundary.
    The resulting per-step relevance cone is accumulated into a
    persistent global BEV relevance map.
    }
    \label{fig:frontier}
\end{figure}

\subsection{System Overview}

\ours{} is built around a \emph{semantic-to-geometric cascade}
(Figure~\ref{fig:pipeline}).
At every step, a dense pixel-level relevance map between $I_t$ and $I_g$
is computed and projected onto a top-down BEV relevance map.
This shared representation serves a dual role: it scores candidate
frontiers for the \textbf{Frontier Exploration} policy, and its peak value
$\srel(t)$ acts as a proximity trigger.
Once $\srel(t)$ exceeds a threshold $\theta$, the agent is
likely in the visual neighborhood of the goal, and \textbf{Geometric
Verification} is invoked on a short-memory cache to test whether reliable
registration is possible.
The geometry model self-certifies its output camera pose for $I_g$ via its correspondence confidence $\sconf$; if $\sconf > \tau$, \textbf{Last-Meter Localization}
uses a wider long-memory cache for precise 6-DoF recovery, and the agent
navigates to the recovered pose, continuously refining it as confidence
improves.
If verification rejects, control returns to Frontier Exploration until the cascade is triggered again.

\subsection{Frontier Exploration}

\paragraph{\textbf{BEV Relevance Map.}}
We show the construction of BEV relevance map in Figure \ref{fig:frontier}.
At each step we extract DINOv2~\cite{oquab2023dinov2} features
$\phi(I_t) \in \mathbb{R}^{H \times W \times d}$ and compute a
per-pixel relevance map against the goal:
\begin{equation}
    \mathbf{S}_t^{(i,j)}
    = \frac{\phi(I_t)^{(i,j)} \cdot \phi(I_g)^{(i,j)}}
           {\|\phi(I_t)^{(i,j)}\|\,\|\phi(I_g)^{(i,j)}\|}, \quad i = 1,\ldots,H,\quad j = 1,\ldots,W.
    \label{eq:simmap}
\end{equation}
To project $\mathbf{S}_t$ onto the BEV occupancy map $\mathcal{M}$,
we compress it to a $W$-dimensional ray vector aligned with the
agent's viewing frustum.
Pixels near the optical center carry the most reliable signal, so
each row is weighted by a Gaussian centered on the image midpoint,
$\mathbf{W}^{(i,j)} \propto
\exp\!\bigl(-\tfrac{(i-H/2)^2}{2\sigma_H^2}\bigr)$,
and the weighted maximum is taken along the height dimension:
\begin{equation}
    \mathbf{s}_t^{(j)}
    = \max_{i}\,\bigl(\mathbf{W} \odot \mathbf{S}_t\bigr)^{(i,j)}.
    \label{eq:gwmax}
\end{equation}
Each ray is further attenuated toward the field-of-view boundary by
an angular mask
\begin{equation}
    m(\varphi) = \cos^2\!\!\left(\frac{\varphi}{\theta_\text{fov}/2}
              \cdot \frac{\pi}{2}\right),
    \label{eq:conemask}
\end{equation}
where $\varphi$ is the angle between the ray and the optical axis.
To prevent relevance from bleeding through obstacles, each ray is
truncated at the maximum observed depth for that column,
$d_t^{(j)} = \max_i D_t^{(i,j)}$.
The per-step BEV relevance map is:
\begin{equation}
    \mathbf{R}_t(p)
    = m(\varphi(p))\,\mathbf{s}_t^{(j(p))}\,
      \mathbf{1}\!\left(\rho(p) \le d_t^{(j(p))}\right),
    \label{eq:bevmap}
\end{equation}
where $\rho(p)$ is the radial distance of BEV cell $p$ from the
camera centre.
$\mathbf{R}_t$ is accumulated into a persistent global map
$\mathbf{G}$ via the weighted-averaging scheme of
VLFM~\cite{yokoyama2024vlfm}.

\paragraph{\textbf{Frontier Selection.}}
Given $\mathbf{G}$, candidate frontiers are scored using four
factors: semantic relevance $S$, geodesic distance $D$, heading
deviation $H$, and information gain $E$ (unexplored pixel ratio
within a fixed radius).

The key design insight is that the relative importance of these
factors must be adaptive: when a frontier shows strong
semantic evidence, the agent should commit to it regardless of
distance or heading cost; when no frontier stands out semantically,
the agent should explore efficiently.
This motivates a gated weighting scheme.
Efficiency factors are normalised via
$\mathrm{Norm}(x) = \frac{1}{2}+\frac{1}{2}
\tanh\!\left(\frac{x-\mathbb{E}(x)}{\mathrm{Std}(x)}\right)$,
which avoids the pathology of min-max normalisation on near-uniform
distributions.
The relevance score is capped as
$S_\text{norm} = \min(S/\theta_\text{relev},\,1)$
so that below-threshold frontiers remain proportionally ranked.
The remaining factors are normalised as
$D_\text{norm} = \mathrm{Norm}(1/D)$,
$H_\text{norm} = \mathrm{Norm}(1 - A/180°)$ where $A$ is the
heading deviation angle,
and $E_\text{norm} = \mathrm{Norm}(E)$.
The composite frontier score is:
\begin{equation}
    \mathrm{score}(f) =
        \underbrace{w_s \cdot S_\text{norm}}_{\text{semantic pull}}
        +
        \underbrace{(1 - S_\text{norm})\bigl(w_d \cdot D_\text{norm}
            + w_h \cdot H_\text{norm}\bigr)}_{\text{efficiency
            (suppressed near goal)}}
        +
        \underbrace{w_e \cdot E_\text{norm}}_{\text{coverage}},
    \label{eq:frontier_score}
\end{equation}
with defaults $w_s\!=\!0.60$, $w_d\!=\!0.55$, $w_h\!=\!0.35$,
$w_e\!=\!0.10$.
The $(1-S_\text{norm})$ coupling is the core mechanism: when
$S_\text{norm}\to 1$ the efficiency terms vanish and the agent
commits to the semantically prominent frontier unconditionally;
when $S_\text{norm}\approx 0$ the score reduces to a pure
efficiency objective.

\paragraph{\textbf{Local Policy.}}
The selected frontier is pursued via an occlusion-aware Fast
Marching Method (FMM)~\cite{sethian1996fast}.
Standard FMM selects the short-term waypoint as the cell minimising
the geodesic distance field, but is agnostic to obstacle clearance,
which can trap the agent in narrow passages.
We augment it with a two-tier clearance mechanism derived from a
local Euclidean Distance Transform (EDT).
A \emph{hard exclusion zone} of radius $r_\text{agent}$
unconditionally removes any candidate cell whose EDT value falls
below $r_\text{agent}$, since the agent cannot physically traverse
a gap narrower than its own body.
A \emph{graded clearance bonus} then reduces the effective distance
cost of remaining candidates in proportion to their obstacle
clearance:
\begin{equation}
    \tilde{d}(c) = d(c) - w_\text{obs} \cdot
    \frac{\min(\mathrm{EDT}(c),\, r_\text{safe})}{r_\text{safe}},
    \label{eq:clearance}
\end{equation}
where $d(c)$ is the FMM geodesic distance and $\tilde{d}(c)$ the
clearance-adjusted cost.
Cells beyond $r_\text{safe}$ receive the full bonus, preventing
the planner from sacrificing goal progress for unnecessary clearance.

\subsection{Geometric Verification}
\label{sec:verification}

Invoking the geometry model at every step is wasteful; registration
is meaningless before any visual overlap with the goal exists.
We gate invocation with the semantic proximity score: $\srel(t) = \max_{i,j}\, \mathbf{S}_t^{(i,j)}$.
The geometry model is invoked only when $\srel(t) > \theta$ for the sake of system efficiency, and $\theta$ is set default as 0.0014 from empirical experience.

When triggered, a short-memory window
$\mathcal{H}_\text{short} = \{I_{t-m+1}, \ldots, I_t\}$ of $m$
recent frames, together with $I_g$, is passed to
VGGT~\cite{wang2025vggt}.
The temporally central frame is placed first in the input sequence
as the reference $I_\text{ref}$: as the frame with maximum average
overlap with all others in the window, it provides the most stable
anchor for relative pose estimation.

Following RobustVGGT~\cite{han2025emergent}, we repurpose VGGT's
final-layer cross-attention activations and $\ell_2$-normalised
feature similarities as proximity confidence signals.
For each frame $I_k \in \mathcal{H}_\text{short} \cup \{I_g\}$, we
compute a raw attention score $r^\text{att}_{k \to \text{ref}}$
(mean cross-attention from $I_k$ to $I_\text{ref}$) and a raw
feature similarity score $r^\text{feat}_{k \to \text{ref}}$ (mean
cosine similarity of final-layer feature maps).
Both score lists are independently min-max normalised across the
$m+1$ frames:
\begin{equation}
    \hat{r}^\text{att}_k =
    \frac{r^\text{att}_{k\to \text{ref}}
          - \min_j r^\text{att}_{j \to \text{ref}}}
         {\max_j r^\text{att}_{j \to \text{ref}}
          - \min_j r^\text{att}_{j \to \text{ref}}},
    \label{eq:minmax}
\end{equation}
and analogously for $\hat{r}^\text{feat}_k$.
The confidence score for the goal image $I_g$ is then:
\begin{equation}
    \sconf = w_\alpha \cdot \hat{r}^\text{att}_g
           + w_f \cdot \hat{r}^\text{feat}_g,
    \quad w_\alpha = w_f = 0.5,
    \label{eq:sconf}
\end{equation}
where $\hat{r}^\text{att}_g$ and $\hat{r}^\text{feat}_g$ denote the
normalised attention and feature scores of $I_g$ specifically.
$\sconf$ is not an external classifier: it is the model's
own internal measure of whether $I_g$ is geometrically-correspondent with the recent observations, repurposed here as a navigation commitment signal.
If $\sconf > \tau$, verification \emph{accepts} and triggers
Last-Meter LocaliZation; otherwise it \emph{rejects} and the cascade
returns to Frontier Exploration.

\subsection{Last-Meter Localization}
\label{sec:localization}

Upon acceptance, the agent switches from the short-memory verification
window to the \emph{long-memory} cache
$\mathcal{H}_\text{long} = \{I_{t-K + 1}, \ldots, I_t\}$ of $K$ frames whose recorded positions ${p_{t-K+1},...p_t}$ span at at least three different positions to enable a well-conditioned Sim(3) alignment.

\paragraph{\textbf{Pose Estimation.}}
We assemble $\mathcal{I} = \mathcal{H}_\text{long} \cup \{I_g\}$ and
pass it to a multi-view 3D foundation model in a single forward pass,
obtaining estimated camera poses $\{\hat{\mathbf{T}}_1, \ldots,
\hat{\mathbf{T}}_K, \hat{\mathbf{T}}_g\}$ in the model's canonical
coordinate frame, where $\hat{\mathbf{T}}_g$ is the estimated pose of
$I_g$. We use Pi3~\cite{wang2025pi} for this stage, as its
architecture handles long-sequence inputs more reliably than
VGGT~\cite{wang2025vggt}.

\paragraph{\textbf{Coordinate Alignment.}}
We align the estimated trajectory to the agent's global coordinate
system in three steps.

\emph{Step 1: Sim(3) position alignment.}
Let $X_k$ and $Y_k$ denote the camera centres extracted from the
predicted and reference (agent-recorded) poses respectively.
We solve for scale $s$, rotation $\mathbf{R}$, and translation
$\mathbf{t}$ via Umeyama's closed-form Sim(3) estimator:
\begin{equation}
    (s^*, \mathbf{R}^*, \mathbf{t}^*) = \arg\min_{s,\mathbf{R},\mathbf{t}}
    \sum_{k=1}^{K} \bigl\|Y_k - (s\,\mathbf{R}\,X_k + \mathbf{t})\bigr\|^2.
    \label{eq:sim3}
\end{equation}

\emph{Step 2: Orientation correction.}
Sim(3) aligns positions but can leave a residual rotation between
predicted and reference frame orientations.
We compute an additional rotation $\mathbf{R}_\text{or}$ that
minimises the mean angular difference between Sim(3)-aligned and
reference rotations.
This correction is applied as a rotation \emph{around the centroid}
$\bar{Y} = \frac{1}{K}\sum_k Y_k$ of the reference cameras,
preventing camera positions from drifting away from the reference
cluster:
\begin{equation}
    X_k^{(3)} \leftarrow
    \mathbf{R}_\text{or}\,(X_k^{(2)} - \bar{Y}) + \bar{Y},
    \label{eq:orient_fix}
\end{equation}
where $X_k^{(2)}$ denotes the position after Step 1.

\emph{Step 3: Apply to the target pose.}
The composed transform $(s^*, \mathbf{R}^*, \mathbf{t}^*,
\mathbf{R}_\text{or})$ is applied identically to $\hat{\mathbf{T}}_g$,
yielding the aligned goal pose
$\mathbf{T}_g^* = (s^*, \mathbf{R}_\text{or}\mathbf{R}^*,
\mathbf{t}^*) \cdot \hat{\mathbf{T}}_g$.
We extract the 2D goal position $\mathbf{p}_g^*$ and heading
$\psi_g^*$ as the navigation target.

\paragraph{\textbf{Confidence-Monitored Refinement.}}
The pose estimate at the moment of first acceptance may carry
residual error if the agent was at the boundary of the goal's
visual neighborhood.
As the agent navigates toward $\mathbf{T}_g^*$, $\sconf$ is
re-evaluated at every step.
Whenever $\sconf$ exceeds its running maximum $\sconf^\text{best}$,
the agent re-invokes localization with the updated long-memory cache:
\begin{equation}
    \mathbf{T}_g^* \;\leftarrow\; \mathbf{T}^*_t \cdot
    \hat{\mathbf{T}}_g^{(t)}
    \quad \text{if } \sconf(t) > \sconf^\text{best}.
    \label{eq:refinement}
\end{equation}
This refinement is motivated by a geometric observation: as the
agent approaches the goal, the photometric overlap between
$\mathcal{H}_\text{long}$ and $I_g$ generally increases, improving
correspondence quality and therefore pose accuracy.
Rather than committing to a fixed estimate at a single decision
point, \ours{} continuously accumulates the best estimate it
encounters during approach, ensuring that the final committed pose
reflects maximum registration confidence.

%% file: subtex/4_experiment.tex
\section{Experiment}
\label{sec:experiment}

\subsection{Experimental Setup}

\textbf{Tasks and datasets.}
We evaluate on two benchmarks in the Habitat~\cite{savva2019habitat}
simulator.
For \textit{ImageNav}, we use the Gibson~\cite{xia2018gibson} dataset
with the 14-scene, 4.2k-episode evaluation split from
Chaplot et al.~\cite{chaplot2020neural}.
For \textit{InstanceImageNav}, we use the HM3D~\cite{yadav2023habitat}
validation set with the 36-scene, 1000-episode split from
Krantz et al.~\cite{krantz2023navigating}.

\textbf{Evaluation metrics.}
We report Success Rate (\textit{SR}) and Success weighted by Path
Length (\textit{SPL})~\cite{anderson2018evaluation} for both tasks.
In addition, we introduce two pose-precision metrics not reported by
prior methods: mean BEV position error $\epsilon_\text{pos}$ and mean
yaw error $\epsilon_\text{head}$ (Section~\ref{sec:method}), computed
at the stop position over all episodes to directly measure
manipulation readiness independent of the $1\,$m proximity threshold. We evaluate these 2 metrics on open-sourced methods.


\begin{figure}[t]
    \centering
    \includegraphics[width=\linewidth]{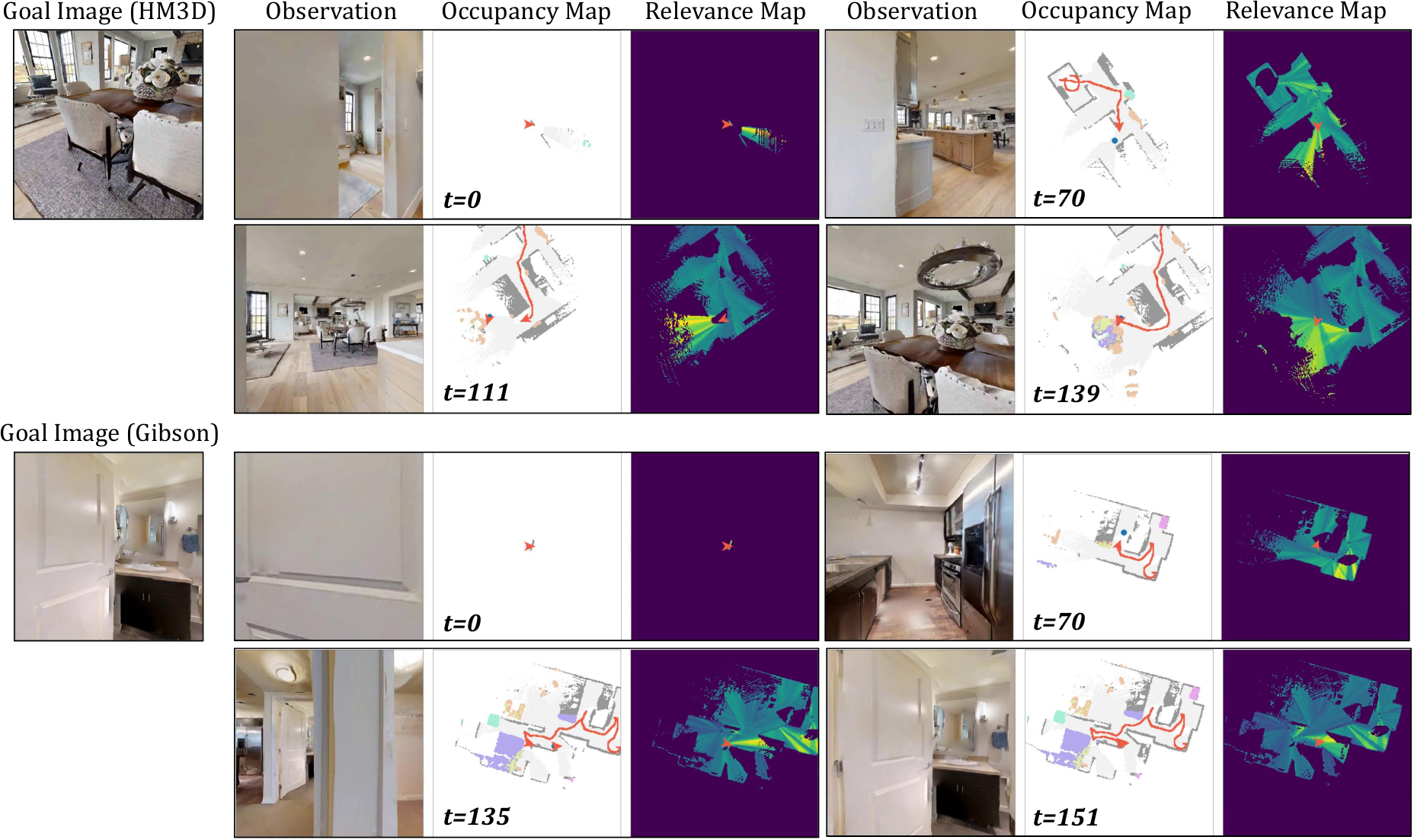}
    \caption{
        \textbf{Navigation examples on HM3D (top) and Gibson (bottom).}
        In both episodes, the BEV relevance map directs the agent toward
goal-relevant regions well before the target is reached (shown at $t=70$).
At $t=111$ (HM3D) and $t=135$ (Gibson), the target pose is
accurately recovered while the agent is still approaching, and the agent
subsequently navigates to the estimated goal position with high precision.
    }
    \label{fig:navigation}
\end{figure}

\begin{table*}[t]
\begin{minipage}[t]{0.5\linewidth}
\makeatletter\def\@captype{table}
\caption{
    \textbf{ImageNav results.}
    $\uparrow$ means higher is better; $\downarrow$ means lower is
    better. \textbf{Bold}: best result; \underline{underline}: second best.
}
\vspace{-8pt}
\centering
\label{tab:main_image}
\resizebox{1.0\linewidth}{!}{
\begin{tabular}{lccccc}
\toprule
\multirow{2}{*}{\textbf{Method}} &
\multirow{2}{*}{\textbf{SR$\uparrow$}} &
\multirow{2}{*}{\textbf{SPL$\uparrow$}} &
\multicolumn{2}{c}{\textbf{Pose Precision}} \\
\cmidrule(lr){4-5}
& & &
$\epsilon_\text{pos}\downarrow$ (m) &
$\epsilon_\text{head}\downarrow$ (°) \\
\midrule
ZER~\cite{al2022zero}
  & 0.292 & 0.216 & 4.84 & 30.02 \\
ZSON~\cite{majumdar2022zson}
  & 0.369 & 0.280 & 3.61 & 26.89 \\
OVRL~\cite{yadav2023offline}
  & 0.542 & 0.270 & - & - \\
OVRL-v2~\cite{yadav2023ovrl}
  & 0.820 & 0.587 & - & - \\
FGPrompt-MF~\cite{sun2023fgprompt}
  & 0.907 & 0.621 & 0.71 & 13.17 \\
FGPrompt-EF~\cite{sun2023fgprompt}
  & 0.904 & \underline{0.665} & 0.75 & 13.21 \\
REGNav~\cite{li2025regnav}
  & \underline{0.929} & \textbf{0.671} & \underline{0.50} & \underline{10.22} \\
\midrule
\ours{}
  & \textbf{0.931} & 0.410 & \textbf{0.27} & \textbf{3.41} \\
\bottomrule
\end{tabular}
}
\end{minipage}
\hfill
\begin{minipage}[t]{0.5\linewidth}
\makeatletter\def\@captype{table}
\caption{
    \textbf{InsImageNav results.}
    $\uparrow$ means higher is better; $\downarrow$ means lower is
    better.
    \textbf{Bold}: best result; \underline{underline}: second best.
}
\label{tab:main_ins}
\vspace{-8pt}
\centering
\resizebox{1.0\linewidth}{!}{
\begin{tabular}{lccccc}
\toprule
\multirow{2}{*}{\textbf{Method}} &
\multirow{2}{*}{\textbf{SR$\uparrow$}} &
\multirow{2}{*}{\textbf{SPL$\uparrow$}} &
\multicolumn{2}{c}{\textbf{Pose Precision}} \\
\cmidrule(lr){4-5}
& & &
$\epsilon_\text{pos}\downarrow$ (m) &
$\epsilon_\text{head}\downarrow$ (°) \\
\midrule
RL Baseline~\cite{krantz2022instance}
  & 0.083 & 0.035 & 6.84 & 47.27 \\
OVRL-v2 IIN~\cite{yadav2023ovrl}
  & 0.248 & 0.118 & - & - \\
Mod-IIN~\cite{krantz2023navigating}
  & 0.561 & 0.233 & - & - \\
UniGoal~\cite{yin2025unigoal}
  & 0.602 & 0.237 & 3.89 & 25.74 \\
IEVE Mask RCNN~\cite{lei2024instance}
  & 0.684 & 0.241 & \underline{2.46} & \underline{20.31} \\
IEVE InternImage~\cite{lei2024instance}
  & 0.702 & 0.252 & - & - \\
GaussNav~\cite{lei2025gaussnav}
  & 0.725 & \underline{0.578} & - & - \\
GauScoreMap~\cite{deng2025hierarchical}
  & \underline{0.784} & \textbf{0.605} & - & - \\
\midrule
\ours{}
  & \textbf{0.826} & 0.259 & \textbf{0.21} & \textbf{1.23} \\
\bottomrule
\end{tabular}
}
\end{minipage}
\end{table*}

\subsection{Main Results}
\textbf{Image Goal Navigation (Gibson).}
Table~\ref{tab:main_image} compares \ours{} against prior learning-based
methods on Gibson.
\ours{} achieves the highest success rate (\textbf{93.1\%}), surpassing
the previous best REGNav~\cite{li2025regnav} by $0.2\%$, and reduces
$\epsilon_\text{pos}$ from 0.50\,m to \textbf{0.27\,m} and
$\epsilon_\text{head}$ from 10.22° to \textbf{3.41°}.
The SPL is lower than the top learning-based methods because Gibson
scenes are compact and some goals are near the start: to accumulate
sufficient long-memory diversity for a well-conditioned Sim(3)
alignment, the agent must travel further before committing, incurring
path overhead on these short episodes.

\textbf{Instance Image Goal Navigation (HM3D).}
Table~\ref{tab:main_ins} compares \ours{} against prior methods on HM3D.
Despite being training-free, \ours{} achieves the highest success rate
(\textbf{82.6\%}), surpassing the previous best
GauScoreMap~\cite{deng2025hierarchical} by $4.2\%$.
Notably, both GauScoreMap and GaussNav~\cite{lei2025gaussnav} achieve
substantially higher SPL by exploiting pre-built Gaussian Splatting
maps that provide shortest-path access to the goal once localized;
\ours{} navigates without any such prior and still outperforms them on
the success rate, demonstrating stronger goal-discovery capability in
unknown environments.
On pose precision, \ours{} achieves \textbf{0.21\,m} and
\textbf{1.23°}, a 10--30$\times$ improvement over methods for which
these metrics can be computed.

We provide two navigation examples on both HM3D and Gibson scenes in the Habitat simulator shown in Figure~\ref{fig:navigation}.

\subsection{Ablation Study}
\label{sec:ablation}

We ablate the key design choices of \ours{} in two groups and evaluate on the HM3D validation set, results are
in Table~\ref{tab:ablation}.

\textbf{Frontier Exploration.}
Every single-factor scoring variant underperforms the full combination:
relevance $S$ alone reaches the goal vicinity but lacks efficiency
pressure; distance $D$ alone ignores goal-directed semantics;
information gain $G$ alone biases toward open space.
The adaptive weighting is therefore necessary for both SR and pose
precision.
Replacing the safety-augmented FMMPlanner with a standard planner
causes a $3.9\%$ SR drop from narrow-passage entrapments.
For the relevance gate threshold $\theta$, lowering it to $0.0010$
gives a marginal SR gain (+$0.2\%$) at the cost of more frequent
(and expensive) geometry-model invocations; raising it to $0.0018$
causes a $6.2\%$ SR drop as valid goal-adjacent views are suppressed.
Replacing DINOv2 with CLIP degrades SR by $6.9\%$, since CLIP's
global image-level features lack the fine-grained details needed for dense per-pixel BEV projection.

\textbf{Geometric Verification and Localization.}
Over-rejection ($\tau\!=\!0.15$) hurts SR more than premature
acceptance ($\tau\!=\!0.05$): $-1.9\%$ vs.\ $-1.1\%$ in SR, reflecting
that a falsely rejected view permanently diverts the agent, whereas a
prematurely accepted estimate can be partially corrected by the
refinement loop.
Both a shorter ($|\mathcal{H}_\text{short}|\!=\!2$) and a longer
($|\mathcal{H}_\text{short}|\!=\!6$) short-memory window degrade SR
($-10.2\%$ and $-6.5\%$ respectively): two frames provide
insufficient viewpoint diversity; six frames dilute temporal locality
and reduce overlap with the goal image.
Removing confidence-monitored refinement causes a $-5.4\%$ SR drop
and substantially worsens pose errors, confirming that the looped confidence-monitored pose refinement is essential for precision.

\begin{table*}[t]
\begin{minipage}[t]{0.65\linewidth}
\makeatletter\def\@captype{table}
\caption{
    \textbf{Ablation study} on HM3D val set.
    Each row removes or replaces one component.
}
\vspace{-8pt}
\centering
\label{tab:ablation}
\resizebox{1.0\linewidth}{!}{
\begin{tabular}{lccccc}
\toprule
\textbf{Configuration} &
\textbf{SR$\uparrow$} &
\textbf{SPL$\uparrow$} &
$\epsilon_\text{pos} (m)\downarrow$ &
$\epsilon_\text{head} (^\circ)\downarrow$ \\
\midrule
Full \ours{}
  & \underline{0.826} & \underline{0.259} & \underline{0.21} & \underline{1.23} \\
\midrule
\multicolumn{5}{l}{\emph{Frontier Exploration ablations}} \\
\quad score with only relevance score $S$
  & 0.786 & 0.231 & 0.42 & 2.07 \\
\quad score with only geodesic distance $D$
  & 0.751 & 0.239 & 0.91 & 2.95 \\
\quad score with only information gain $G$
  & 0.749 & 0.244 & 1.13 & 1.98 \\
\quad w/o safe FMMPlanner (plain FMMPlanner)
  & 0.787 & 0.233 & 0.54 & 2.16 \\
\quad relevance threshold $\theta=0.0010$
  & \textbf{0.828} & \textbf{0.261} & \textbf{0.20} & \textbf{1.19} \\
\quad relevance threshold $\theta=0.0018$
  & 0.764 & 0.221 & 1.01 & 1.78 \\
\quad replace DINOv2~\cite{oquab2023dinov2} with CLIP~\cite{radford2021learning} visual encoder
  & 0.757 & 0.217 & 1.08 & 2.96 \\
\midrule
\multicolumn{5}{l}{\emph{Geometric Verification and Localization ablations}} \\
\quad confidence threshold $\tau=0.05$
  & 0.815 & 0.254 & 0.68 & 1.37 \\
\quad confidence threshold $\tau=0.15$
  & 0.807 & 0.250 & 1.01 & 1.88 \\
\quad short memory length $|\mathcal{H}_\text{short}|=2$
  & 0.724 & 0.221 & 1.76 & 3.01 \\
\quad short memory length $|\mathcal{H}_\text{short}|=6$
  & 0.761 & 0.227 & 1.31 & 2.24 \\
\quad w/o Confidence-Monitored Refinement
  & 0.772 & 0.230 & 1.25 & 2.65 \\
\bottomrule
\end{tabular}
}
\end{minipage}
\hfill
\begin{minipage}[t]{0.34\linewidth}
\makeatletter\def\@captype{table}
\centering
\caption{\textbf{Failure case statistics} on all 1000 HM3D validation episodes.}
\label{tab:failure}
\vspace{-8pt}
\resizebox{1.0\linewidth}{!}{
    \begin{tabular}{lcc}
    \toprule
    \textbf{Failure reason} &
    \textbf{$N_{fail}$} &
    \textbf{$R_{fail}$} \\
    \midrule
    Wrong pose estimation & 55 & 31.6\% \\
    Goal not found & 46 & 26.4\% \\
    Agent stuck & 34 & 19.5\% \\
    Premature acceptance & 26 & 15.0\% \\
    Correct region rejected & 13 & 7.5\% \\
    \midrule
    Total & 174 & 100\% \\
    \bottomrule
    \end{tabular}
}
\end{minipage}
\end{table*}

\subsection{Failure Case Analysis}
We analyze the 174 failed episodes on the HM3D validation set and
categorize them into five failure modes, reported in
Table~\ref{tab:failure}.

\textbf{Wrong pose estimation (31.6\%)} is the most frequent failure
mode. These episodes reach the correct vicinity and pass geometric
verification, but the recovered 6-DoF pose is inaccurate. This
occurs either because the long-memory frames provide insufficient
viewpoint diversity for a well-conditioned Sim(3) alignment, or
because the foundation model produces unreliable pose estimates.

\textbf{Goal not found (26.4\%)} covers episodes where neither the
semantic proximity threshold $\theta$ nor the confidence threshold
$\tau$ is exceeded within the episode step budget. This 
occurs in large-scale scenes where the agent exhausts its steps
traversing or revisiting explored regions but does not find semantically-relevant or geometrically-confident regions.

\textbf{Agent stuck (19.5\%)} refers to episodes where the agent
enters a narrow passage or dead-end from which the local planner
cannot recover. Although the safety-augmented FMMPlanner
substantially reduces this failure mode relative to the standard
planner (Table~\ref{tab:ablation}), a residual fraction persists
in environments with particularly tight geometry where the hard
exclusion zone radius is insufficient to prevent entrapment.

\textbf{Premature acceptance (15.0\%)} occurs when $\sconf$ exceeds
$\tau$ while the agent is still outside the true goal neighborhood.
The foundation model establishes spurious correspondences between
the goal image and a visually similar but geometrically incorrect
location, most commonly in environments with repeated structural
patterns such as symmetric furniture layouts or identically decorated
rooms.

\textbf{Correct region rejected (7.5\%)} is the complementary
failure: the agent reaches the true goal vicinity but $\sconf$
remains below $\tau$.
This affects goals in structurally confined
spaces such as toilets and plants in walls, where
the geometry model finds too few reliable correspondences to exceed
the confidence threshold.

%% file: subtex/5_discussion.tex
\section{Discussion}
\label{sec:discussion}

\ours{} advances image goal navigation toward manipulation-ready
precision while remaining training-free.
We discuss three limitations and the directions they motivate.

\textbf{Dependence on depth and odometry.}
The system relies on depth and odometry sensors for BEV map construction.
Streaming multi-view foundation
models~\cite{zhuo2025streaming,lan2025stream3r,chen2025ttt3r,yuan2026infinitevggt}
could in principle enable robotic agents free from depth and odometry sensors with purely RGB-derived trajectories,
but we have tested them, only to find they all suffer from temporal pose drift over the episode lengths typical of navigation tasks.
Consequently, developing a robust and efficient 3D foundation model remains an open
challenge for sensor-light image-goal navigation.

\textbf{Exploration under sparse semantic signal.}
The 26.4\% of failures due to \emph{Goal not found}
reflects a fundamental limitation: when no semantically prominent
frontier exists, the scoring function degrades to a pure efficiency
objective with no directional bias toward the goal.
Incorporating scene-level priors such as room-layout estimates or object co-occurrence statistics could
provide a useful signal during the semantically silent phase of exploration.

\textbf{Fixed thresholds.}
Roughly 22\% of failures stem from $\tau$ being either too permissive
(15.0\%) or too conservative (7.5\%).
While our default hyperparameters generalize well across both
benchmarks without per-scene tuning, replacing fixed thresholds with
adaptive counterparts calibrated to scene texture density or learned
from navigation experience could further balance the two complementary failure modes.

%% file: subtex/6_conclusion.tex
\section{Conclusion}
\label{sec:conclusion}

We presented \ours{}, a training-free system that advances Image Goal
Navigation from coarse proximity to pose-precise localization.
By treating the goal image as an unposed geometric query and
registering it against the agent's accumulated observations via a
3D multi-view foundation model, our method recovers the exact
camera pose of the goal, a capability that prior semantic matching
methods cannot provide.
The semantic-to-geometric cascade ensures efficiency by invoking the
geometry model only when meaningful visual overlap exists, with no
scene reconstruction, task-specific training or explicit semantic label prediction required.
\ours{} achieves state-of-the-art success rates on both benchmarks:
93.1\% on Gibson and 82.6\% on HM3D, while reducing position error
to 0.27\,m and heading error to 3.41° on Gibson, and 0.21\,m and
1.23° on HM3D.
These results suggest that pose-accurate visual localization can
serve as a reliable bridge between image goal navigation and
downstream manipulation tasks, enabling robots to act on
image-specified goals rather than merely arrive near them.

%% file: main.bib
@String(ICLR  = {Int. Conf. Learn. Represent.})

@String(AAAI  = {AAAI})

@String(ICLR  = {ICLR})

@inproceedings{lei2024instance,
  title={Instance-aware exploration-verification-exploitation for instance imagegoal navigation},
  author={Lei, Xiaohan and Wang, Min and Zhou, Wengang and Li, Li and Li, Houqiang},
  booktitle={Proceedings of the IEEE/CVF Conference on Computer Vision and Pattern Recognition},
  pages={16329--16339},
  year={2024}
}

@INPROCEEDINGS{7989381,
  author={Zhu, Yuke and Mottaghi, Roozbeh and Kolve, Eric and Lim, Joseph J. and Gupta, Abhinav and Fei-Fei, Li and Farhadi, Ali},
  booktitle={2017 IEEE International Conference on Robotics and Automation (ICRA)}, 
  title={Target-driven visual navigation in indoor scenes using deep reinforcement learning}, 
  year={2017},
  volume={},
  number={},
  pages={3357-3364},
  doi={10.1109/ICRA.2017.7989381}}

@article{krantz2022instance,
  title={Instance-specific image goal navigation: Training embodied agents to find object instances},
  author={Krantz, Jacob and Lee, Stefan and Malik, Jitendra and Batra, Dhruv and Chaplot, Devendra Singh},
  journal={arXiv preprint arXiv:2211.15876},
  year={2022}
}

@inproceedings{krantz2023navigating,
  title={Navigating to objects specified by images},
  author={Krantz, Jacob and Gervet, Theophile and Yadav, Karmesh and Wang, Austin and Paxton, Chris and Mottaghi, Roozbeh and Batra, Dhruv and Malik, Jitendra and Lee, Stefan and Chaplot, Devendra Singh},
  booktitle={Proceedings of the IEEE/CVF International Conference on Computer Vision},
  pages={10916--10925},
  year={2023}
}

@article{deng2025hierarchical,
  title={Hierarchical Scoring with 3D Gaussian Splatting for Instance Image-Goal Navigation},
  author={Deng, Yijie and Yuan, Shuaihang and Bethala, Geeta Chandra Raju and Tzes, Anthony and Liu, Yu-Shen and Fang, Yi},
  journal={arXiv preprint arXiv:2506.07338},
  year={2025}
}

@article{lei2025gaussnav,
  title={Gaussnav: Gaussian splatting for visual navigation},
  author={Lei, Xiaohan and Wang, Min and Zhou, Wengang and Li, Houqiang},
  journal={IEEE Transactions on Pattern Analysis and Machine Intelligence},
  volume={47},
  number={5},
  pages={4108--4121},
  year={2025},
  publisher={IEEE}
}

@inproceedings{yin2025unigoal,
  title={Unigoal: Towards universal zero-shot goal-oriented navigation},
  author={Yin, Hang and Xu, Xiuwei and Zhao, Linqing and Wang, Ziwei and Zhou, Jie and Lu, Jiwen},
  booktitle={Proceedings of the IEEE/CVF Conference on Computer Vision and Pattern Recognition},
  pages={19057--19066},
  year={2025}
}

@article{narasimhan2025splatsearch,
  title={SplatSearch: Instance Image Goal Navigation for Mobile Robots using 3D Gaussian Splatting and Diffusion Models},
  author={Narasimhan, Siddarth and Lisondra, Matthew and Wang, Haitong and Nejat, Goldie},
  journal={arXiv preprint arXiv:2511.12972},
  year={2025}
}

@inproceedings{al2022zero,
  title={Zero experience required: Plug \& play modular transfer learning for semantic visual navigation},
  author={Al-Halah, Ziad and Ramakrishnan, Santhosh Kumar and Grauman, Kristen},
  booktitle={Proceedings of the IEEE/CVF Conference on Computer Vision and Pattern Recognition},
  pages={17031--17041},
  year={2022}
}

@inproceedings{chaplot2020neural,
  title={Neural topological slam for visual navigation},
  author={Chaplot, Devendra Singh and Salakhutdinov, Ruslan and Gupta, Abhinav and Gupta, Saurabh},
  booktitle={Proceedings of the IEEE/CVF conference on computer vision and pattern recognition},
  pages={12875--12884},
  year={2020}
}

@article{majumdar2022zson,
  title={Zson: Zero-shot object-goal navigation using multimodal goal embeddings},
  author={Majumdar, Arjun and Aggarwal, Gunjan and Devnani, Bhavika and Hoffman, Judy and Batra, Dhruv},
  journal={Advances in Neural Information Processing Systems},
  volume={35},
  pages={32340--32352},
  year={2022}
}

@inproceedings{yadav2023offline,
  title={Offline visual representation learning for embodied navigation},
  author={Yadav, Karmesh and Ramrakhya, Ram and Majumdar, Arjun and Berges, Vincent-Pierre and Kuhar, Sachit and Batra, Dhruv and Baevski, Alexei and Maksymets, Oleksandr},
  booktitle={Workshop on Reincarnating Reinforcement Learning at ICLR 2023},
  year={2023}
}

@article{sun2023fgprompt,
  title={FGPrompt: Fine-grained goal prompting for image-goal navigation},
  author={Sun, Xinyu and Chen, Peihao and Fan, Jugang and Chen, Jian and Li, Thomas and Tan, Mingkui},
  journal={Advances in Neural Information Processing Systems},
  volume={36},
  pages={12054--12073},
  year={2023}
}

@inproceedings{wang2025vggt,
  title={Vggt: Visual geometry grounded transformer},
  author={Wang, Jianyuan and Chen, Minghao and Karaev, Nikita and Vedaldi, Andrea and Rupprecht, Christian and Novotny, David},
  booktitle={Proceedings of the Computer Vision and Pattern Recognition Conference},
  pages={5294--5306},
  year={2025}
}

@article{wang2025pi,
  title={$\pi^3$: Permutation-Equivariant Visual Geometry Learning},
  author={Wang, Yifan and Zhou, Jianjun and Zhu, Haoyi and Chang, Wenzheng and Zhou, Yang and Li, Zizun and Chen, Junyi and Pang, Jiangmiao and Shen, Chunhua and He, Tong},
  journal={arXiv preprint arXiv:2507.13347},
  year={2025}
}

@article{lin2025depth,
  title={Depth anything 3: Recovering the visual space from any views},
  author={Lin, Haotong and Chen, Sili and Liew, Junhao and Chen, Donny Y and Li, Zhenyu and Shi, Guang and Feng, Jiashi and Kang, Bingyi},
  journal={arXiv preprint arXiv:2511.10647},
  year={2025}
}

@article{han2025emergent,
  title={Emergent Outlier View Rejection in Visual Geometry Grounded Transformers},
  author={Han, Jisang and Hong, Sunghwan and Jung, Jaewoo and Jang, Wooseok and An, Honggyu and Wang, Qianqian and Kim, Seungryong and Feng, Chen},
  journal={arXiv preprint arXiv:2512.04012},
  year={2025}
}

@article{oquab2023dinov2,
  title={Dinov2: Learning robust visual features without supervision},
  author={Oquab, Maxime and Darcet, Timoth{\'e}e and Moutakanni, Th{\'e}o and Vo, Huy and Szafraniec, Marc and Khalidov, Vasil and Fernandez, Pierre and Haziza, Daniel and Massa, Francisco and El-Nouby, Alaaeldin and others},
  journal={arXiv preprint arXiv:2304.07193},
  year={2023}
}

@inproceedings{yokoyama2024vlfm,
  title={Vlfm: Vision-language frontier maps for zero-shot semantic navigation},
  author={Yokoyama, Naoki and Ha, Sehoon and Batra, Dhruv and Wang, Jiuguang and Bucher, Bernadette},
  booktitle={2024 IEEE International Conference on Robotics and Automation (ICRA)},
  pages={42--48},
  year={2024},
  organization={IEEE}
}

@inproceedings{savva2019habitat,
  title={Habitat: A platform for embodied ai research},
  author={Savva, Manolis and Kadian, Abhishek and Maksymets, Oleksandr and Zhao, Yili and Wijmans, Erik and Jain, Bhavana and Straub, Julian and Liu, Jia and Koltun, Vladlen and Malik, Jitendra and others},
  booktitle={Proceedings of the IEEE/CVF international conference on computer vision},
  pages={9339--9347},
  year={2019}
}

@article{yadav2023ovrl,
  title={Ovrl-v2: A simple state-of-art baseline for imagenav and objectnav},
  author={Yadav, Karmesh and Majumdar, Arjun and Ramrakhya, Ram and Yokoyama, Naoki and Baevski, Alexei and Kira, Zsolt and Maksymets, Oleksandr and Batra, Dhruv},
  journal={arXiv preprint arXiv:2303.07798},
  year={2023}
}

@inproceedings{li2025regnav,
  title={Regnav: Room expert guided image-goal navigation},
  author={Li, Pengna and Wu, Kangyi and Fu, Jingwen and Zhou, Sanping},
  booktitle={Proceedings of the AAAI Conference on Artificial Intelligence},
  volume={39},
  number={5},
  pages={4860--4868},
  year={2025}
}

@inproceedings{xia2018gibson,
  title={Gibson env: Real-world perception for embodied agents},
  author={Xia, Fei and Zamir, Amir R and He, Zhiyang and Sax, Alexander and Malik, Jitendra and Savarese, Silvio},
  booktitle={Proceedings of the IEEE conference on computer vision and pattern recognition},
  pages={9068--9079},
  year={2018}
}

@inproceedings{yadav2023habitat,
  title={Habitat-matterport 3d semantics dataset},
  author={Yadav, Karmesh and Ramrakhya, Ram and Ramakrishnan, Santhosh Kumar and Gervet, Theo and Turner, John and Gokaslan, Aaron and Maestre, Noah and Chang, Angel Xuan and Batra, Dhruv and Savva, Manolis and others},
  booktitle={Proceedings of the IEEE/CVF Conference on Computer Vision and Pattern Recognition},
  pages={4927--4936},
  year={2023}
}

@article{sethian1996fast,
  title={A fast marching level set method for monotonically advancing fronts.},
  author={Sethian, James A},
  journal={proceedings of the National Academy of Sciences},
  volume={93},
  number={4},
  pages={1591--1595},
  year={1996}
}

@inproceedings{radford2021learning,
  title={Learning transferable visual models from natural language supervision},
  author={Radford, Alec and Kim, Jong Wook and Hallacy, Chris and Ramesh, Aditya and Goh, Gabriel and Agarwal, Sandhini and Sastry, Girish and Askell, Amanda and Mishkin, Pamela and Clark, Jack and others},
  booktitle={International conference on machine learning},
  pages={8748--8763},
  year={2021},
  organization={PmLR}
}

@article{anderson2018evaluation,
  title={On evaluation of embodied navigation agents},
  author={Anderson, Peter and Chang, Angel and Chaplot, Devendra Singh and Dosovitskiy, Alexey and Gupta, Saurabh and Koltun, Vladlen and Kosecka, Jana and Malik, Jitendra and Mottaghi, Roozbeh and Savva, Manolis and others},
  journal={arXiv preprint arXiv:1807.06757},
  year={2018}
}

@inproceedings{zhu2017target,
  title={Target-driven visual navigation in indoor scenes using deep reinforcement learning},
  author={Zhu, Yuke and Mottaghi, Roozbeh and Kolve, Eric and Lim, Joseph J and Gupta, Abhinav and Fei-Fei, Li and Farhadi, Ali},
  booktitle={2017 IEEE international conference on robotics and automation (ICRA)},
  pages={3357--3364},
  year={2017},
  organization={ieee}
}

@inproceedings{kim2023topological,
  title={Topological semantic graph memory for image-goal navigation},
  author={Kim, Nuri and Kwon, Obin and Yoo, Hwiyeon and Choi, Yunho and Park, Jeongho and Oh, Songhwai},
  booktitle={Conference on Robot Learning},
  pages={393--402},
  year={2023},
  organization={PMLR}
}

@inproceedings{kwon2023renderable,
  title={Renderable neural radiance map for visual navigation},
  author={Kwon, Obin and Park, Jeongho and Oh, Songhwai},
  booktitle={Proceedings of the IEEE/CVF Conference on Computer Vision and Pattern Recognition},
  pages={9099--9108},
  year={2023}
}

@article{savinov2018semi,
  title={Semi-parametric topological memory for navigation},
  author={Savinov, Nikolay and Dosovitskiy, Alexey and Koltun, Vladlen},
  journal={arXiv preprint arXiv:1803.00653},
  year={2018}
}

@inproceedings{wasserman2023last,
  title={Last-mile embodied visual navigation},
  author={Wasserman, Justin and Yadav, Karmesh and Chowdhary, Girish and Gupta, Abhinav and Jain, Unnat},
  booktitle={Conference on Robot Learning},
  pages={666--678},
  year={2023},
  organization={PMLR}
}

@inproceedings{guo2025igl,
  title={IGL-Nav: Incremental 3D Gaussian Localization for Image-goal Navigation},
  author={Guo, Wenxuan and Xu, Xiuwei and Yin, Hang and Wang, Ziwei and Feng, Jianjiang and Zhou, Jie and Lu, Jiwen},
  booktitle={Proceedings of the IEEE/CVF International Conference on Computer Vision},
  pages={6808--6817},
  year={2025}
}

@article{kerbl20233d,
  title={3d gaussian splatting for real-time radiance field rendering.},
  author={Kerbl, Bernhard and Kopanas, Georgios and Leimk{\"u}hler, Thomas and Drettakis, George and others},
  journal={ACM Trans. Graph.},
  volume={42},
  number={4},
  pages={139--1},
  year={2023}
}

@inproceedings{arandjelovic2016netvlad,
  title={NetVLAD: CNN architecture for weakly supervised place recognition},
  author={Arandjelovic, Relja and Gronat, Petr and Torii, Akihiko and Pajdla, Tomas and Sivic, Josef},
  booktitle={Proceedings of the IEEE conference on computer vision and pattern recognition},
  pages={5297--5307},
  year={2016}
}

@inproceedings{sattler2018benchmarking,
  title={Benchmarking 6dof outdoor visual localization in changing conditions},
  author={Sattler, Torsten and Maddern, Will and Toft, Carl and Torii, Akihiko and Hammarstrand, Lars and Stenborg, Erik and Safari, Daniel and Okutomi, Masatoshi and Pollefeys, Marc and Sivic, Josef and others},
  booktitle={Proceedings of the IEEE conference on computer vision and pattern recognition},
  pages={8601--8610},
  year={2018}
}

@inproceedings{sarlin2019coarse,
  title={From coarse to fine: Robust hierarchical localization at large scale},
  author={Sarlin, Paul-Edouard and Cadena, Cesar and Siegwart, Roland and Dymczyk, Marcin},
  booktitle={Proceedings of the IEEE/CVF conference on computer vision and pattern recognition},
  pages={12716--12725},
  year={2019}
}

@inproceedings{sarlin2020superglue,
  title={Superglue: Learning feature matching with graph neural networks},
  author={Sarlin, Paul-Edouard and DeTone, Daniel and Malisiewicz, Tomasz and Rabinovich, Andrew},
  booktitle={Proceedings of the IEEE/CVF conference on computer vision and pattern recognition},
  pages={4938--4947},
  year={2020}
}

@inproceedings{sun2021loftr,
  title={LoFTR: Detector-free local feature matching with transformers},
  author={Sun, Jiaming and Shen, Zehong and Wang, Yuang and Bao, Hujun and Zhou, Xiaowei},
  booktitle={Proceedings of the IEEE/CVF conference on computer vision and pattern recognition},
  pages={8922--8931},
  year={2021}
}

@inproceedings{wang2024dust3r,
  title={Dust3r: Geometric 3d vision made easy},
  author={Wang, Shuzhe and Leroy, Vincent and Cabon, Yohann and Chidlovskii, Boris and Revaud, Jerome},
  booktitle={Proceedings of the IEEE/CVF conference on computer vision and pattern recognition},
  pages={20697--20709},
  year={2024}
}

@article{zhuo2025streaming,
  title={Streaming 4d visual geometry transformer},
  author={Zhuo, Dong and Zheng, Wenzhao and Guo, Jiahe and Wu, Yuqi and Zhou, Jie and Lu, Jiwen},
  journal={arXiv preprint arXiv:2507.11539},
  year={2025}
}

@article{lan2025stream3r,
  title={Stream3r: Scalable sequential 3d reconstruction with causal transformer},
  author={Lan, Yushi and Luo, Yihang and Hong, Fangzhou and Zhou, Shangchen and Chen, Honghua and Lyu, Zhaoyang and Yang, Shuai and Dai, Bo and Loy, Chen Change and Pan, Xingang},
  journal={arXiv preprint arXiv:2508.10893},
  year={2025}
}

@article{yuan2026infinitevggt,
  title={InfiniteVGGT: Visual Geometry Grounded Transformer for Endless Streams},
  author={Yuan, Shuai and Yang, Yantai and Yang, Xiaotian and Zhang, Xupeng and Zhao, Zhonghao and Zhang, Lingming and Zhang, Zhipeng},
  journal={arXiv preprint arXiv:2601.02281},
  year={2026}
}

@article{chen2025ttt3r,
  title={Ttt3r: 3d reconstruction as test-time training},
  author={Chen, Xingyu and Chen, Yue and Xiu, Yuliang and Geiger, Andreas and Chen, Anpei},
  journal={arXiv preprint arXiv:2509.26645},
  year={2025}
}
